# A Factorized Variational Technique for Phase Unwrapping in Markov Random Fields

**Kannan Achan,**　　**Brendan J. Frey**
Adaptive Algorithms Laboratory
University of Toronto
http://www.cs.toronto.edu/~frey/aal

**Ralf Koetter**
Coordinated Sciences Laboratory
Electrical and Computer Engineering
University of Illinois at Urbana

## Abstract

Some types of medical and topographic imaging device produce images in which the pixel values are "phase-wrapped", *i.e.*, measured modulus a known scalar. *Phase unwrapping* can be viewed as the problem of inferring the integer number of relative shifts between each and every pair of neighboring pixels, subject to an *a priori* preference for smooth surfaces, and a zero curl constraint, which requires that the shifts must sum to 0 around every loop. We formulate phase unwrapping as a probabilistic inference problem in a Markov random field where the prior favors the zero curl constraint. We derive a relaxed, factorized variational method that infers approximations to the marginal probabilities of the integer shifts between pairs of neighboring pixels. The original, unwrapped image can then be obtained by integrating the integer shifts. We compare our mean field technique with the least squares method on a synthetic 100 × 100 image, and give results on a larger 512 × 512 image measured using synthetic aperature radar from Sandia National Laboratories.

## 1 INTRODUCTION

Phase unwrapping is an easily stated, fundamental problem in signal processing [1]. The signal is measured modulus a known wavelength, which we take to be 1 without loss of generality. Fig. 1b shows the wrapped, 1-dimensional signal obtained from the original signal shown in Fig. 1a. The objective of phase unwrapping is to estimate the original signal from the wrapped version, using knowledge about which signals are more probable *a priori*. Without prior knowledge, the wrapped signal itself provides an error-free guess

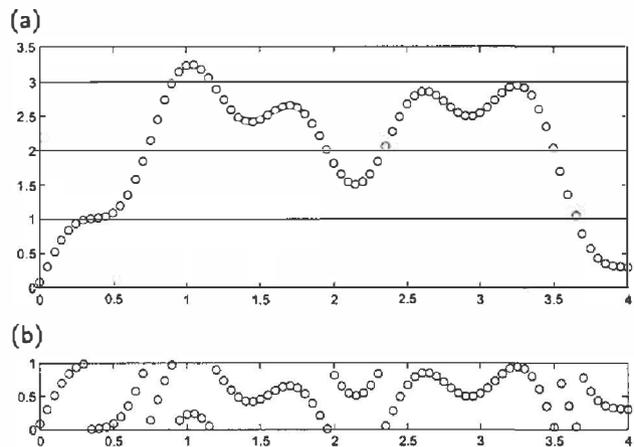

Figure 1: (a) A 1-dimensional signal. (b) The phase-wrapped version of the signal in (a), where the wavelength is 1.

at the unwrapped image.

Two-dimensional phase unwrapping has applications in a variety of sensory modalities, including magnetic resonance imaging [2] and interferometric synthetic aperture radar (SAR) [3]. Fig. 2a) shows a phase-wrapped image produced by MRI, and Fig. 2b shows a phase-wrapped image representing phase-wrapped topographic height measurements of terrain from Sandia National Laboratories. It turns out that 2-dimensional phase unwrapping is a much more difficult problem than 1-dimensional phase unwrapping.

A sensible goal in phase unwrapping is to infer the number of relative wrappings, or integer "shifts", between every pair of neighboring measurements. Positive shifts correspond to an increase in the number of wrappings in the direction of the $x$ or $y$ coordinate, whereas negative shifts correspond to a decrease in the number of wrappings in the direction of the $x$ or $y$ coordinate. Once the relative shifts are known, we can arbitrarily assign an absolute number of wrappings to one point and determine the absolute number of wrap-pings at any other point by summing the shifts along a path connecting the two points. To account for direction, when taking a step against the direction of the coordinate, the shift should be subtracted. Integrating the shifts in this fashion for all points produces an unwrapped signal.

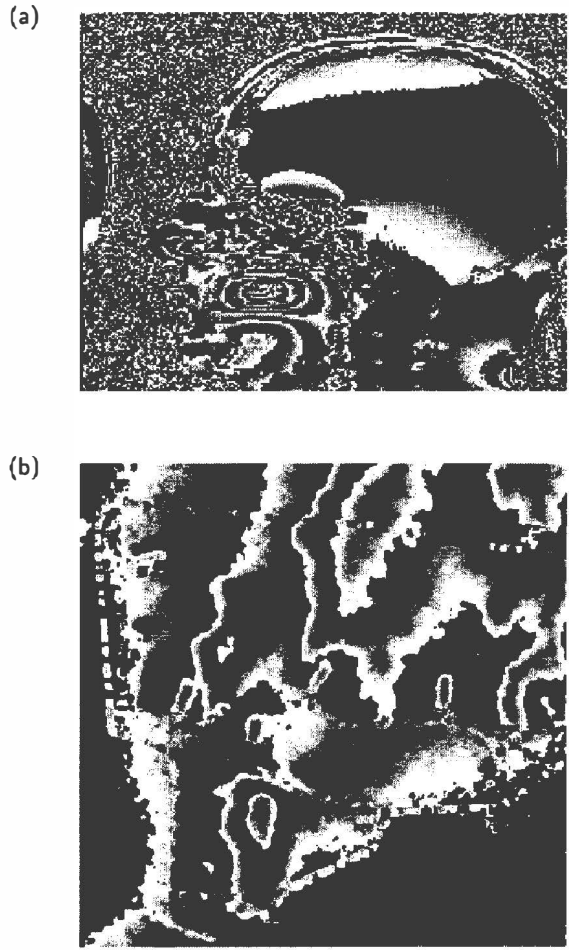

(a)

(b)

Figure 2: Phase-wrapped images from (a) magnetic resonance imaging data (courtesy of Z.-P. Liang) and (b) synthetic aperture radar data (courtesy of Sandia National Laboratories, New Mexico). Pixel values close to 0 are painted white, whereas pixel values close to 1 (the wrapping wavelength) are painted black.

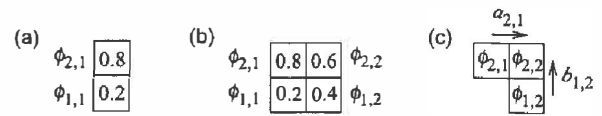

Figure 3: Phase measurements in small image patches. From (a) it appears that a shift occurred between points $1,1$ and $2,1$. From (b), it appears that a shift probably did *not* occur between points $1,1$ and $2,1$. (c) The phase at $2,2$ can be predicted from the phases at $2,1$ and $1,2$, plus the shifts $a_{2,1}$ and $b_{1,2}$.

When neighboring signal values are more likely closer together than further apart *a priori*, 1-dimensional signals can be unwrapped optimally in time that is linear in the signal length. For every pair of neighboring measurements, the shift that makes the unwrapped values as close together as possible is chosen. For example, the shift between 0.4 and 0.5 would be 0, whereas the shift between 0.9 and 0.0 would be $-1$.

For 2-dimensional signals, there are many possible 1-dimensional paths between any two points. These paths should be examined in combination, since the sum of the shifts along every such path should be equal. Viewing the shifts as state variables, the cut-set between any two points is exponential in the size of the grid, making exact inference NP-hard [4].

Previous approaches to solving the phase unwrapping problem include least squares estimates (these are *not* MMSE estimates) [5, 3, 6, 2], integer programming methods [7, 4] and branch cut techniques [8].

Our approach to phase unwrapping is to construct a probability model on the shift variables and then use an approximate probabilistic inference technique to infer the shifts. Only a subset of the possible configurations of the shift variables will lead to valid gradient field. In particular, the sum of the shifts around any loop in the image must be zero. We refer to this constraint as the *zero curl constraint*.

We formulate phase unwrapping as factorized variational inference (mean field inference) problem in an relaxed probability model, where the prior favors shifts that satisfy the zero curl constraint. We relax the prior by introducing a temperature parameters. The preference for shifts that satisfy the zero curl constraint is weakened at high temperatures. As the temperature is decreased to zero (annealing), the model settles to a consistent configuration of the shifts.

## 2   RELAXED GRADIENT FIELD MODEL

Let $\phi_{i,j} \in [0,1)$ be the phase value at $i,j$. (We assume that measurements are taken *modulus* 1 – *i.e.*, the wavelength is 1.) Let $a_{i,j} \in \mathcal{I}$ be the unknown shift between points $i,j$ and $i,j+1$. So, the difference in the unwrapped values at pixels $i,j+1$ and $i,j$ is $\phi_{i,j+1} - \phi_{i,j} - a_{i,j}$. Similarly, let $b_{i,j} \in \mathcal{I}$ be the unknown shift between points at $i,j$ and $i+1,j$.

Consider the two patches of image shown in Fig. 3. From Fig. 3a, the difference in the unwrapped values at $1,1$ and $2,1$ is $0.8 - 0.2 - a_{1,1}$. Assuming the values are more likely to be closer together than further apart, we decide that $a_{1,1} = 1$, so that $0.8 - 0.2 - a_{1,1}$ is as close to 0 as possible.

We can make these local decisions for every neighboring pair of points in a large image, but the resulting set of shifts will not satisfy the constraint of summing to zero around every loop. If we make local decisions for the patch in Fig. 3b, then we decide that $a_{1,1} = 1$, $b_{1,2} = 0$, $a_{2,1} = 0$ and $b_{1,1} = 0$. The sum of these shifts around a counter-clockwise loop is $a_{1,1} + b_{1,2} - a_{2,1} - b_{1,1} = 1$, giving a *curl violation*. We can fix this curl violation by changing one or more of the shifts, at the cost of not keeping the unwrapped pixel differences as close to zero as possible.

Notice that if the the sum of the shifts around every $2 \times 2$ loop is zero, then the sum of the shifts around any loop is zero. So, the $2 \times 2$ loops provide a sufficient set of constraints.

To choose the form of the above cost, we develop a probability model of the shifts and the observed phases. We choose a prior for the shifts that favors shifts that satisfy the zero curl constraint:

$$p(a,b) \propto \prod_{i,j} \exp[-(a_{i,j} + b_{i,j+1} - a_{i,j+1} - b_{i,j})^2/T]$$

To incorporate a preference for smooth surfaces, we restrict the values of the $a$'s and $b$'s to be in $\{-1, 0, 1\}$.

The density of the observed phase measurements can be formulated recursively. As shown in Fig. 3c, the phase at 2,2 can be predicted from the phases at 2,1 and 1,2, plus the shifts $a_{2,1}$ and $b_{1,2}$. The prediction from 2,1 is $\phi_{2,1} + a_{2,1}$, while the prediction from 1,2 is $\phi_{1,2} + b_{1,2}$. The average prediction is $(\phi_{2,1} + a_{2,1} + \phi_{1,2} + b_{1,2})/2$. Assuming a Gaussian likelihood we obtain the general form

$$p(\phi|a,b) \propto \prod_{i,j} \Big( \exp[-(\phi_{i,j+1} - \phi_{i,j} - a_{i,j})^2/2\sigma^2] \\ \cdot \exp[-(\phi_{i+1,j} - \phi_{i,j} - b_{i,j})^2/2\sigma^2] \Big)$$

The joint distribution $p(a,b,\phi) = p(a,b)p(\phi|a,b)$ is

$$p(a,b,\phi) \propto \prod_{i,j} \exp[-(a_{i,j} + b_{i,j+1} - a_{i,j+1} - b_{i,j})^2/T] \\ \cdot \prod_{i,j} \Big( \exp[-(\phi_{i,j+1} - \phi_{i,j} - a_{i,j})^2/2\sigma^2] \\ \cdot \exp[-(\phi_{i+1,j} - \phi_{i,j} - b_{i,j})^2/2\sigma^2] \Big)$$

### 2.1 Temperature

The temperature $T$ allows the prior to be relaxed. For $T \to \infty$, the zero curl constraint is completely relaxed. For $T \to 0$, only shifts that satisfy the zero curl constraint have nonvanishing probability.

## 3 FACTORIZED VARIATIONAL INFERENCE

Exact inference (*e.g.*, computing $p(a_{i,j}|\phi)$) in the above model is intractable. So, we use a factorized variational technique, a.k.a. a mean field approximation (see [9]).

We approximate $p(a,b|\phi)$ with a factorized distribution,

$$q(a,b) = \prod_{i,j} q(a_{i,j})q(b_{i,j}). \tag{1}$$

We parameterize the $q$-distribution as follows:

$$q(a_{i,j} = k) = \alpha_{i,j,k}, \quad q(b_{i,j} = k) = \beta_{i,j,k}, \tag{2}$$

where we require $\sum_{k=-1}^{1} \alpha_{i,j,k} = 1$ and so on. The $\alpha$'s and $\beta$'s are variational parameters.

To bring $q$ "close" to $p$, we would like to minimize the relative entropy,

$$D = \sum_{a,b} q(a,b) \log \frac{q(a,b)}{p(a,b|\phi)}. \tag{3}$$

However, this quantity contains $p(a,b|\phi)$ for which we do not have a simple, closed form expression.

Subtracting $\log p(\phi)$ (which does not depend on the variational parameters) from the above relative entropy, we obtain a cost function that *can* be easily minimized:

$$F = D - \log p(\phi) \\ = \sum_{a,b} q(a,b) \log \frac{q(a,b)}{p(a,b,\phi)} \\ = \ldots \\ = \sum_{i,j} \Big( \sum_{k=-1}^{1} \alpha_{i,j,k} \log \alpha_{i,j,k} + \sum_{k=-1}^{1} \beta_{i,j,k} \log \beta_{i,j,k} \Big) \\ + \frac{1}{T} \sum_{i,j,k,l,m,n} \alpha_{i,j,k} \beta_{i,j+1,l} \alpha_{i+1,j,m} \beta_{i,j,n}(k+l-m-n)^2 \\ + \frac{1}{2\sigma^2} \sum_{i,j} \Big( \sum_{k=-1}^{1} \alpha_{i,j,k} (\phi_{i,j+1} - \phi_{i,j} - k)^2 \\ + \sum_{k=-1}^{1} \beta_{i,j,k} (\phi_{i+1,j} - \phi_{i,j} - k)^2 \Big). \tag{4}$$

For the results presented below, we use a conjugate gradient optimizer (including Langrangian constraints to ensure that $\sum_{k=-1}^{1} \alpha_{i,j,k} = 1$ and so on).

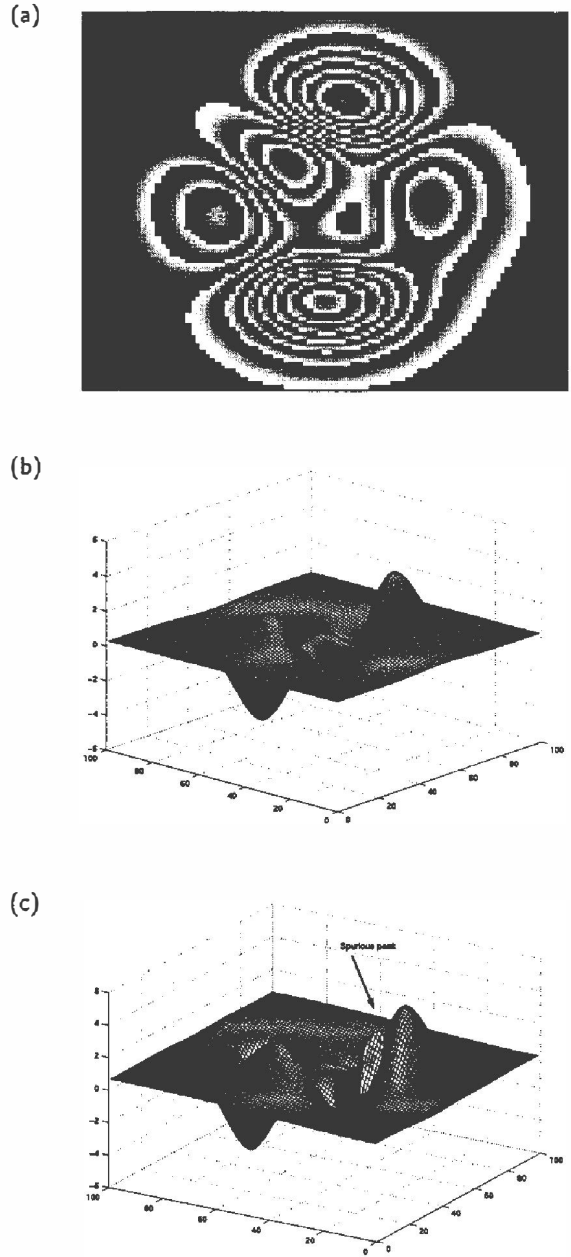

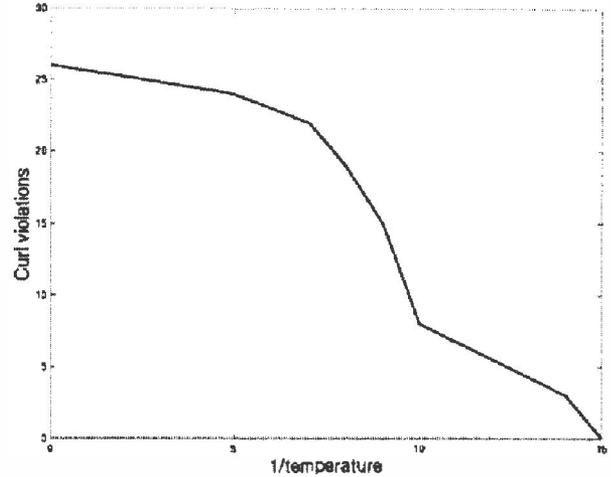

Figure 5: Number of curl violations as a function of inverse temperature.

### 4.1 Synthetic Data

Fig. 4a shows the phase-wrapped image produced from our synthetic data. After minimizing $F$ using 20 iterations of conjugate gradients, while annealing the temperature from a high value to a low value, we obtained a set of shift probabilities ($\alpha$'s and $\beta$'s from above). For each pair of pixels, we picked the shift that had highest probability. The resulting set of shifts satisfied the zero curl constraint. From the shifts, we obtained a gradient field and integrated it to obtain the surface shown in Fig. 4b. This surface matches the original surface perfectly.

We applied the least squares method to the wrapped data in Fig. 4a and obtained the surface shown in Fig. 4b. In contrast to our variational method, the least squares method produces ridge-like artifacts.

Fig. 5 shows the number of zero curl violations as a function of the inverse temperature. In this experiment we ran the conjugate gradient optimizer to convergence (about 5 iterations) for each temperature. There appears to be a critical point at an inverse temperature of about 8.

Figure 4: (a) A 100 × 100 wrapped image. (b) Unwrapped surface produced by our variational technique. (c) Unwrapped surface produced by the least squares method.

## 4  EXPERIMENTAL RESULTS

We present results on two images. In the first case, we synthesized the original image (surface), so we know the "ground truth" and can easily compare our method with the standard least squares technique [5, 3, 6, 2]. In the second case, we present results on unwrapping the Sandia image.

### 4.2 Data from Sandia National Laboratories

After minimizing $F$ using as input the 512×512 phase-wrapped image from the Sandia National Laboratories, New Mexico (Fig. 2b), we found that there were still some zero curl violations. Using the shifts to produce a "gradient field" produces a "gradient field" that violates the zero curl constraint. So, we used our method as a preprocessor for the least squares method, obtaining the surface shown in Fig. 6. When the least squares method is used directly (without using our method as a preprocessor), the least squares method produces a surface that has greater error than our method, when compared to the wrapped input data.

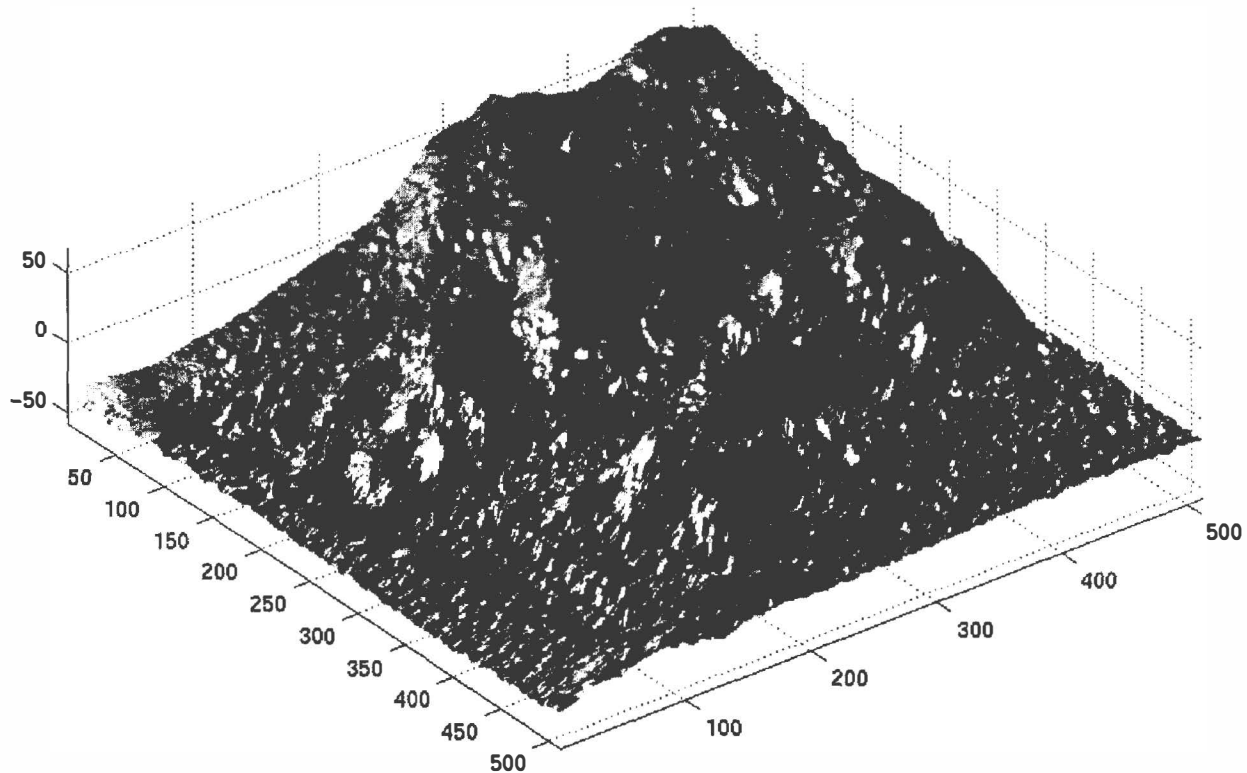

Figure 6: Unwrapped surface produced by our variational method applied to the 512 × 512 Sandia data shown in Fig. 2b.

## 5　MEASURE OF UNCERTAINITY

An interesting and useful consequence of obtaining the variational Q distribution is that it allows for the computation of a measure to address uncertainty in selecting the most probable shift. The parameters of the Q distribution ($\alpha$'s and $\beta$'s) that correspond to every pair of adjacent pixels represent the posterior probability of the posible shifts between them. This can be used to compute the entropy,

$$E(i,j) = -\sum_k \alpha_{i,j,k} \log \alpha_{i,j,k} \qquad (5)$$

Entropy is maximum when the probabilities are equal, implying inability to favour a particular shift. The notion of minimizing uncertainty is embedded in our cost function F. The first term relates to entropy of the Q distribution and hence a competing goal in the optimization process is the assignment of disticnt probabilities to the individual shifts.

We have shown the results of unwrapping the $100 \times 100$ surface (Fig. 4), along with the entropy in Fig. 7. It can be observed that entropy is very high initially and gradually decreases as the annealing proceeds.

## 6　SUMMARY

We formulated phase unwrapping in 2-dimensional topologies as a factorized variational inference problem in a relaxed Markov random field. The method outperforms the least squares method on a synthetic $100 \times 100$ surface. We also showed that the method can unwrap more realistic sizes of image, such as the $512 \times 512$ topographic map from Sanida National Laboratories.

As with most annealing methods, the algorithm exhibits critical point behavior. We are currently performing experments to gain insight into the critical points.

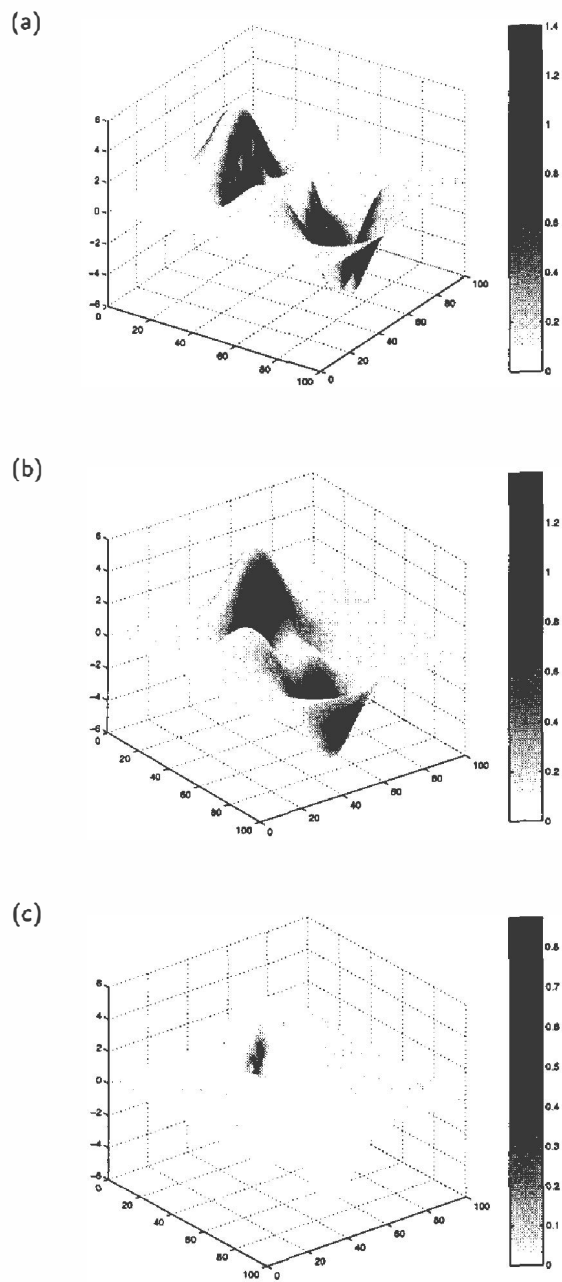

Figure 7: (a) 100 × 100 unwrapped image after 2 iterations. Black indicates high entropy(refer to color bar). The surface shown was generated by ignoring curl violations. (b)after 4 iterations; there were no curl violations. (c)after 8 iterations.

## References

[1] D. C. Ghiglia and M. D. Pritt, *Two-Dimensional Phase Unwrapping. Theory, Algorithms and Software*, John Wiley & Sons, 1998.

[2] Z.-P. Liang and P. C. Lauterbur, *Principles of Magnetic Resonance Imaging : A Signal Processing Perspective*, IEEE Press, New York NY., 2000.

[3] C. V. Jakowatz, Jr., D. E. Wahl, P. H. Eichel, and P. A. Thompson, *Spotlight-mode Synthetic Aperture Radar: A Signal Processing Approach*, Kluwer Academic Publishers, Boston MA., 1996.

[4] C. W. Chen and H. A. Zebker, "Network approaches to two-dimensional phase unwrapping: intractability and two new algorithms," *Journal of the Optical Society of America A*, vol. 17, no. 3, pp. 401–414, 2000.

[5] S. M.-H. Song, S. Napel, N. J. Pelc, and G. H. Glover, "Phase unwrapping of MR phase images using Poisson's equation," *IEEE Transactions on Image Processing*, vol. 4, no. 5, pp. 667–676, May 1995.

[6] M. Costantini, A. Farina, and F. Zirilli, "A fast phase unwrapping algorithm for SAR interferometry," *IEEE Transactions on Geoscience and Remote Sensing*, vol. 37, no. 1, pp. 452–460, January 1999.

[7] M. Costantini, "A phase unwrapping method based on network programming," in *Proceedings of the Fring '96 Workshop*. 1996, ESA SP-406.

[8] R. M. Goldstein, H. A. Zebker, and C. L. Werner, "Satellite radar interferometry: Two-dimensional phase unwrapping," *Radio Science*, vol. 23, no. 4, pp. 713–720, 1988.

[9] M. I. Jordan, Z. Ghahramani, T. S. Jaakkola, and L. K. Saul, "An introduction to variational methods for graphical models," in *Learning in Graphical Models*, M. I. Jordan, Ed. Kluwer Academic Publishers, Norwell MA., 1998.

## 7 ACKNOWLEDGEMENTS

We thank N. Petrovic and D. Munson for helpful discussions.